\pdfoutput=1

\documentclass[11pt]{article}

\usepackage{acl}
\usepackage{times}
\usepackage{latexsym}
\usepackage{float}
\usepackage{nicefrac}
\usepackage[T1]{fontenc}

\usepackage[utf8]{inputenc}

\usepackage{microtype}
\usepackage{graphicx}
\usepackage{booktabs}
\usepackage{tabularx}
\usepackage{amsmath}
\usepackage{multirow}
\usepackage{subfigure}
\usepackage{bm}
\usepackage{tikz}
\usepackage{url}
\usepackage{amssymb}
\usepackage{xurl}

\newcommand{\tokenex}{\texttt{TokenEx}} 
\newcommand{\tokenintex}{\texttt{TokenIntEx}} 
\newcommand{\spanintex}{\texttt{SpanIntEx}} 

\title{Evaluating Input Feature Explanations through a Unified Diagnostic Evaluation Framework}

\author{Jingyi Sun\quad Pepa Atanasova\quad Isabelle Augenstein
\\University of Copenhagen\\
$\texttt{\{jisu, pepa, augenstein\}@di.ku.dk}$
}

\begin{document}
\maketitle
\begin{abstract}
Explaining the decision-making process of machine learning models is crucial for ensuring their reliability and transparency for end users. One popular explanation form highlights key input features, such as i) tokens (e.g., Shapley Values and Integrated Gradients), ii) interactions between tokens (e.g., Bivariate Shapley and Attention-based methods), or iii) interactions between spans of the input (e.g., Louvain Span Interactions).
However, these explanation types have only been studied in isolation, making it difficult to judge their respective applicability. To bridge this gap, we develop a unified framework that facilitates an automated and direct comparison between highlight and interactive explanations comprised of four diagnostic properties\footnote{Our code can be found at https://github.com/copenlu/A-unified-framework-for-input-feature-explanations}.  
We conduct an extensive analysis across these three types of input feature explanations--each utilizing three different explanation techniques--across two datasets and two models, and reveal that each explanation has distinct strengths across the different diagnostic properties. 
Nevertheless, interactive span explanations outperform other types of input feature explanations across most diagnostic properties. Despite being relatively understudied, our analysis underscores the need for further research to improve methods generating these explanation types. Additionally, integrating them with other explanation types that perform better in certain characteristics could further enhance their overall effectiveness.

\end{abstract}
\section{Introduction}
\begin{figure}
\small
\includegraphics[width=0.51\textwidth]{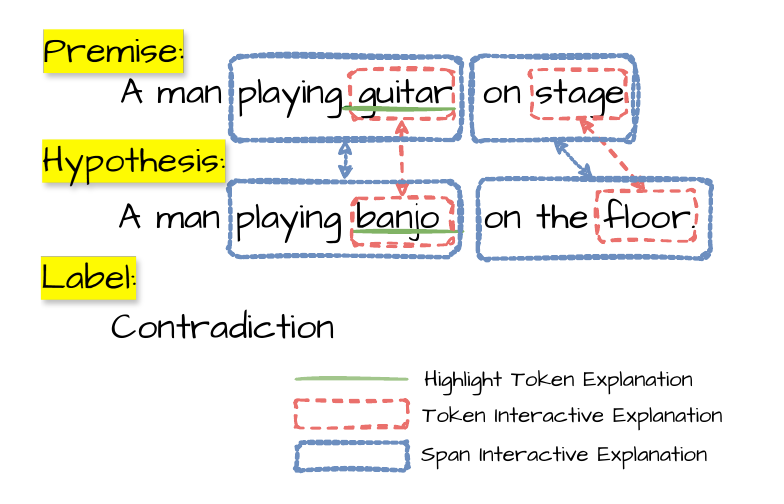}
\caption{\label{p1}An example of the three types of Input Feature Explanations on an instance from the SNLI dataset, with their two most important pieces of explanation (token for \tokenex, token tuple for \tokenintex, span tuple for \spanintex, correspondingly).
}
\end{figure}
Input feature explanations reveal how a model makes decisions based on a specific input. Among these, the most widely used explanation type is \textit{Token Explanations} (\tokenex), which for Natural language Understanding tasks provide importance scores for the tokens of the input, using methods such as Shapley Values \cite{lundberg2017unified} and Integrated Gradients~\citep{sundararajan2017axiomatic}. However, for complex reasoning tasks that require reasoning across multiple pieces of text, e.g., Fact Checking is performed given a claim and evidence, Natural Language Inference is performed given a premise and a hypothesis, \tokenex\ can be insufficient to capture the relations employed between the different parts of the input. To this end, \textit{Token Interactive Explanations} (\tokenintex) are proposed as another explanation type that provides importance scores for interactions between two tokens of the input with methods such as Bivariate Shapley~\cite{masoomi2023explanations} and Layer-wise Attention Attribution~\citep{ye-etal-2021-connecting}. Further, to enhance the semantic coherence of these interactive explanations, \textit{Span Interactive Explanations} (\spanintex) is an explanation type that provides importance scores for interactions across tuples of spans, e.g., generated by Louvain community detection~\citep{choudhury-etal-2023-explaining}. Figure \ref{p1} 
 showcases these three types of input feature explanations.

It is crucial to develop automated, rigorous and comprehensive evaluation frameworks to ensure the principled selection of the most suitable explanation in a practical application \cite{haeun2024naia} and systematic progress in the development of different types of explanations \cite{diagnostic_guided, jolly2022generating}. However, these three types of input feature explanations have typically been studied in isolation, where explanation methods of the same type are compared \cite{atanasova-etal-2020-diagnostic,deyoung-etal-2020-eraser,janizek2021explaining,choudhury-etal-2023-explaining}. 
Moreover, evaluations of interactive explanations have been restricted to one property.
To address these gaps, we develop \textit{a unified framework that facilitates an automated and direct comparison between highlight and interactive input feature explanations on a suite of four diagnostic properties}. The framework allows for a rapid, scalable comparison across explainability methods, essential for the growing number of new techniques. Using the framework, we then perform an \textit{extensive empirical analysis of the properties of existing explanation methods across all three explanation types}.

\textbf{Unified Evaluation Framework.} Our unified evaluation framework consists of four essential diagnostic properties -- Faithfulness, Agreement with Human Annotation, Simulatability, and Complexity. They are the most widely used for \tokenex\ \cite{nauta2023anecdotal} and include the only property used for interactive explanations -- Faithfulness. \textit{Faithfulness} \citep{chen-etal-2020-generating-hierarchical,chen-etal-2021-explaining,choudhury-etal-2023-explaining} measures the extent to which explanations accurately reflect the reasons used by the model in its predictions. \textit{Agreement with Human Annotation} \citep{atanasova-etal-2020-diagnostic} evaluates whether explanations exhibit an inductive bias akin to human reasoning, potentially enhancing their plausibility to end users. \textit{Simulatablity} \citep{pruthi-etal-2022-evaluating} estimates whether the explanations are useful to an agent for replicating the model's decisions. Finally, \textit{Complexity} \citep{10.5555/3491440.3491857} evaluates whether the explanations are easy to comprehend. 
In the unified evaluation framework, we further extend the four properties to facilitate their application and comparison across all three explanation types (\S\ref{sec:methodology:overall}).

\textbf{Extensive Empirical Analysis of Input Feature Explanations.} We conduct an extensive analysis across two different textual tasks, two language models, and three explanation techniques for each input feature explanation type. Our findings indicate that \tokenex\ exhibit greater Comprehensiveness, and \spanintex\ -- Sufficiency.
Additionally, \spanintex\ and \tokenintex\ align more closely with human annotations at the token level than \tokenex. Moreover, \spanintex\ demonstrate the highest interaction overlap with human annotations. Further, \spanintex\ are found to be most beneficial for Simulatability. Finally, our results suggest that \spanintex\ and \tokenex\ are easier to comprehend.

Overall, our analysis highlights the strengths of each explanation type across various diagnostic properties, with \spanintex\ generally outperforming others on most measures. However, we observe a trade-off between Comprehensiveness and plausibility, particularly with \spanintex, underscoring the need for methods that enhance both. Future research could explore integrated approaches that combine explanation types to optimize performance across all diagnostic properties.

\section{A Unified Automated Evaluation Framework for Highlight Explanations}
\label{sec:methodology:overall}
According to the established taxonomy of evaluation approaches for explainability methods, evaluations can be functionally grounded, human-grounded, or application-grounded~\citep{doshi2017towards}. Our work focuses on automated functionally-grounded evaluation, which serves as a foundation for the other evaluation types as it enables the rapid, scalable comparison across explainability methods, essential for the growing number of new techniques. To this end, we present a unified framework comprised of four widely employed diagnostic properties: Faithfulness, Agreement with Human Annotations, Simulatability, and Complexity. This section formally introduces them and outlines the extensions that allow for their application across different input feature explanation types. 
\subsection{Preliminaries}
\label{preliminaries} 
We start with a dataset $D$, and a model $M$ fine-tuned on $D$. An instance $x \in D$ comprises two parts, e.g., a claim and an evidence, the first consisting of $m$ tokens, and the second -- of $n$ tokens.
We apply an explanation attribution method $A_{E}$ of type $E\in \{\tokenex, \tokenintex, \spanintex\}$ to $M$, and each $x \in D$: $A_{E}(M, x) = \{e^x_k, a^x_k | k \in [0, s-1]\}$, where $e^x_k$ is a token/pair of tokens/pair of token spans from the input and $a^x_k$ denotes its importance score. $k$ is the importance ranking of the corresponding piece of explanation. $s$ is an upper limit for the number of most important pieces of explanation for an instance, such that: $s \in [1,m+n]$ for \tokenex, $s \in [1, m \cdot n]$ for \tokenintex, and for \spanintex\ $s$ varies for each instance with $s \in [1, f], f < m! \cdot n!$. Depending on the explanation type $E$, $e^x_k$ can consist of:
one token $x_i$ for \tokenex, $i \in [0, m+n-1]$,
one token pair $(x_p, x_q)$ for \tokenintex, where $p \in [0, m-1]$ and $q \in [m, m+n-1]$,
one span pair $((x_s, \dots, x_{s+l_1}), (x_t, \dots, x_{t+l_2}))$ for \spanintex, where $s, s+l_1 \in [0, m-1]$ and $t, t+l_2 \in [m, m+n-1]$.

Considering a particular threshold for the number of most important explanation pieces $s$, we compute the total set of input tokens involved in the presented explanation for $x$:
\begin{equation}
\small
    T_{A_E,M}(x,s)=\{x_i | x_i \in {e}_{k}^{x}, k \in [0, s-1] \}
\label{threshold_0}
\end{equation}
However, as noted above, the upper bound for $s$ can vary across input feature explanations. Additionally, the number of tokens included in the top-$k$ important explanations can differ substantially among explanation types -- in top-1 we can have: 1 token for \tokenintex, 2 tokens for \tokenintex, and more than 2 tokens for \spanintex. This variability makes it difficult to compare results across different explanation types. To this end, we propose extensions for each of the studied diagnostic properties that result in \textit{unified diagnostic properties} applicable for a direct comparison of the different types of input feature explanations.

\subsection{Faithfulness}
\label{sec:faithfulness:method}
Faithfulness \cite{deyoung-etal-2020-eraser} assesses whether explanations accurately reflect the model’s decision-making process. It involves two aspects -- Sufficiency and Comprehensiveness -- measured as the number of the model's prediction changes after keeping (SP) or omitting (CP) $k$ most important portions of the input (see omission details in \S\ref{sec:app:faithfulness-mask-details}): 
\begin{equation}
\small
    CP(x,A_E,k)=\left\{
    \begin{array}{ll}
    0, & \text{if } f(x-T_{A_E,M}(x,k))=f(x) \\
    1, & \text{otherwise }
    \end{array}
    \right\}
\label{eq:comp-1}
\end{equation}
\begin{equation}
\small
    SP(X,A_E,k)=\left\{
    \begin{array}{ll}
    1, & \text{if } f(T_{A_E,M}(x,k))=f(x) \\
    0, & \text{otherwise }
    \end{array}
    \right\}
\label{eq:comp-3}
\end{equation}
To take different thresholds $k$, we average over $k \in [0, s=m+n-1]$. We then also average the results across instances within $D$ to compute the final Comprehensiveness and Sufficiency scores:
\begin{equation}
\small
Comp_{A_E}= \frac{\sum_{x\in{D}}^{|D|} \sum_{k=0}^{s} CP(x,A_E,k)}{s*|D|}
\label{eq:comp-5}
\end{equation}
\begin{equation}
\small
Suff_{A_E} = \frac{\sum_{x\in{D}}^{|D|} \sum_{k=0}^{s} SP(x,A_E,k)}{s*|D|}
\label{eq:comp-6}
\end{equation}

\textbf{Unified Faithfulness.} To facilitate the comparison of faithfulness scores, we introduce a \textit{dynamic threshold} $\theta_{x,k}$. It represents the number of unique tokens used for a perturbation on $x$, same across explanation methods of all explanation types. Since top-$k$ explanations for $A_E$ of type \spanintex\ typically contain more tokens than for \tokenex\ or \tokenintex, we set $\theta_{x,k}$ across the explanation methods of all types to:
\begin{equation}
\small
\theta_{x,k} = \left|T_{A_{\spanintex},M}(x,k)\right|
\label{eq:threshold_1}
\end{equation}
Thus, $\theta_{x,k}$ becomes a dynamic threshold that adapts based on each instance’s top-$k$ explanation tokens from $A_{\spanintex}$. We then adjust the number of top-$k$ explanations selected from $A_{\tokenex}$ and $A_{\tokenintex}$, $k_{A_{\tokenex}}$ and $k_{A_{\tokenintex}}$, correspondingly, to result in the same number of perturbed tokens $\theta_{x,k}$:
\begin{equation}
\small
\left|T_{A_{\tokenex},M}(x,k_{A_{\tokenex}}(x))\right| = \theta_{x,k}
\label{eq:threshold_2}
\end{equation}
\begin{equation}
\small
\left|T_{A_{\tokenintex},M}(x,k_{A_{\tokenintex}}(x))\right| = \theta_{x,k}
\label{eq:threshold_3}
\end{equation}
Furthermore, a \textit{Random baseline} is established, where the tokens for perturbation are selected randomly to match the average $\theta_{x,k}$ across $D$.

\subsection{Agreement with Human Annotations}
\label{sec:agreement:method}
Agreement with Human Annotations has been used to assess the overlap between generated and human-annotated explanations, which can indicate the plausibility of the generated explanations to end users. For $E=\tokenex$, the measure is computed by calculating a precision score for the top-$k$ most important explanations compared to the gold human annotations~\cite{atanasova-etal-2020-diagnostic}. 

For $E=\tokenex$, $a_i^x, i \in [1,m+n]$ is the attribution score of $i$th most important explanation for $x$. $s=m+n$ is the number of explanations extracted from $x$. Corresponding to each explanation's attribution score, $s$ thresholds are set, forming the threshold list $\omega_{A_E}(x)=[a^x_0,...,a^x_s]$.
By selecting explanations with higher attribution scores than each threshold in $\omega_{A_E}(x)$, $s$ targeted explanation sets are obtained, where $C_{A_E}(x,i)\{e^x_j:a^x_j<=a^x_i\}$ represents the set for the $i^{th}$ threshold, $a^x_j$ is the attribution score of token $e^x_j$ for $E=\tokenex$.

Comparing these sets with the golden explanation set $e^G$, $s$ precision-recall pairs $P_{i}/R_{i}(x,e^G, A_E)$ can be derived. Average Precision (AP) is then obtained by weighting $P_i$ with the corresponding $R_i$ increase:
\begin{equation}
\small
P_{i}/R_{i}(x,e^G, A_E)=Pre/Rec(C_{A_E}(x,i),e^G)
\end{equation}
\begin{equation}\textstyle
\small
AP_{A_E}(x, e^G)=\sum_{i=0}^{s} (R_{i}-R_{i-1})*P_{i}
\end{equation}
Mean AP (MAP) is calculated for all $x\!\in\!D$:
\begin{equation}
\small
MAP_{A_E}=\frac{\sum_{x\in D}^{|D|} AP_{A_E}(x, e^G)}{|D|}
\label{eq:map}
\end{equation}
\textbf{Unified Agreement with Human Annotation Measure.} 
For a fair comparison between the different types of explanations, the thresholds $\omega_{A_E}(x)$ for including the same number of tokens across the explanation methods follows the procedure set for the Unified Faithfulness (\S\ref{sec:faithfulness:method}). 

Furthermore, we measure Agreement with Human Annotations at the \textbf{interaction level} for the gold $\spanintex$/$\tokenintex$ explanations and at the \textbf{token level} for gold $\tokenex$ explanations.   

\textbf{Interaction-level Agreement with Human Annotations.} For a fair comparison between $\tokenintex$ and $\spanintex$ methods, we adapt $MAP_{\tokenex}$ to the interaction level. Specifically, we compute the mean average precision (Eq. \ref{eq:map}) w.r.t. the human-annotated $\tokenintex$/$\spanintex$ sets.

\textbf{Token-level Agreement with Human Annotations.} For a fair comparison between $\tokenex$ and $\tokenintex$/$\spanintex$ methods, we extract tokens from $\tokenintex$/$\spanintex$ and compare them with tokens from golden $\tokenintex$/$\spanintex$ sets. 
To compute MAP at token level, we follow the similar procedure set for $E=\tokenex$ (Eq. \ref{eq:map}) with threshold lists $\omega_{A_{\tokenintex/\spanintex}}(x)$, but the targeted sets $C_{A_{E_{token}}}(x, i)$ contain tokens extracted from $\tokenintex$/$\spanintex$ methods. The golden set $S_{A_{E_{token}}}(x)$ aggregates tokens from golden $\tokenintex$/$\spanintex$ sets.

We also set a \textit{Random baseline}, where the number of randomly selected span pairs, token pairs, or tokens for each instance matches the average number of tokens per instance extracted with a \spanintex\ explanation method.

\subsection{Simulatability}
\label{sec:simulatability:method}
Simulatability was initially proposed to measure how accurately humans can predict a model's outputs based on its explanations \cite{chen2023models,hase-etal-2020-leakage}. Previous studies demonstrated that Simulatability can be approximated using an automated agent model as a surrogate for human understanding \cite{pruthi-etal-2022-evaluating}. \textit{Given the established positive correlation between Simulatability and human evaluation of explanation utility}, we integrate the Simulatability scores obtained from an agent model with different explanation types to approximate their utility for humans.

Following existing work \cite{hase-etal-2020-leakage}, we train an agent model $AM$, sharing the same architecture as the original model $M$, to simulate $M$'s predictions $Y'$ using produced explanations. During $AM$'s training phase, we extract the top-$k$ explanations and incorporate them in the input. 
In comparison, another agent model, $AM_O$, is trained without explanation guidance as a baseline on the same training set.
During the testing phase, the simulation accuracy of $AM$ and $AM_O$ over the shared dataset $D$ is calculated.\footnote{While existing work \cite{hase-etal-2020-leakage} notes that incorporating natural language explanations in the testing phase could leak the predicted label, we use only input feature explanations that do not contain additional information.} The difference between the accuracies is interpreted as the explanation's effect in enhancing the simulatability of $M$:
\begin{equation}
\small
    Sim=ACC(AM(D),Y')-ACC(AM_O(D),Y')
\end{equation}

\paragraph{Unified Simulatability.} To compare the simulation utility of different explanation types, we train a separate agent model $AM_E$ for each explanation method $A_E$ and calculate the corresponding simulation performance on the common test set.
For a fair comparison across the different explanation method types $A_E$, we first ensure top-$k_{E}$ explanations are presented for assisting the agent's training for $A_E$, following Section \ref{sec:faithfulness:method}. This ensures each model is exposed to the same quantity of unique tokens from different explanation types.

During the training of $AM$, we introduce the explanations from $A_E$ into the learning of $AM_{A_E}$; we supplement $x$ with top-$k_{A_E}$ explanations instead so that the agent model is trained with the same mechanism whether the explanations are provided or not, and each training instance will contain the input sequence $x_{A_E}$ and golden label $Y'$ which is predicted by the original model $M$. Specifically, we examine two different ways of presenting the explanations as part of the original input sequence, $I_{Symbol}$ and $I_{Text}$ (see \S\ref{sec:app:detailed-form}.); all aim to ensure the explanations of different types are inserted similarly. 

At test time, the F1 scores of agent models $AM_{E}$ and $AM_{O}$ over $D$ are calculated: 
\begin{equation}
\small
    SF_{E}=F1(AM_{E}(x_{E}),Y'),x \in D
\end{equation}
\begin{equation}
\small
    SF_{O}=F1(AM_{O}(x),Y'),x \in D
\end{equation}
Note that the explanations from $A_E$ are also provided to the input for the unseen instances for agent model $AM_{E}$. The final simulation metric then indicates how much this specific type of input feature explanation enhances the model's simulatability:
\begin{equation}
\small
    RSF_{E}=SF_{E}-SF_{O}
\end{equation}

\subsection{Complexity}
Feature attribution explanations are designed to aid human understanding of a model's reasoning over specific instances. However, since humans have a limited capacity to process large amounts of information simultaneously, these explanations need to be easy to comprehend. Even if we select only the top-$k$ attributions with the highest importance scores, they need to be distinctive as opposed to the attribution scores having a uniform distribution. \citet{10.5555/3491440.3491857} propose to measure the complexity of a produced explanation with entropy \cite{renyi1961measures} over the attribution scores of all the produced explanations by $A_{\tokenex}$ method:
\begin{equation}\textstyle
\small
        P(x,p) =\left|a^x_p\right| / \sum_{q=1}^{m+n}\left|a^x_q\right|
\end{equation}
\begin{equation}\textstyle
\small
        CL(x) =- \sum_{p=1}^{m+n}
        P(x,p)
        \ln(P(x,p))
\end{equation}
$m+n$ is the total number of generated $A_{\tokenex}$ explanations, and all explanations are considered for the complexity score computation. Higher entropy means different features have similar attribution scores, where the simplest explanation, with low entropy, would be concentrated on one
feature.

\textbf{Unified Complexity.} 
To ensure a fair comparison across different types of explanation methods $A_E$, we maintain consistency in the size of the chosen explanation lists across all $A_E$ for the same instance, denoted as $k_{x}$, as the number of generated $A_{\tokenex}$/$A_{\tokenintex}$/$A_{\spanintex}$ explanations originally vary for the same $x$. The complexity score $CL_{A_E}(x,k_{x})$ of the top-$k_{x}$ explanation list under method $A_E$ is calculated as:
\begin{equation}\textstyle
\small
        P_{A_E}(x,k_x,i) =\left| a^x_i\right| /\sum_{j=1}^{k_{x}}\left|a^x_j\right|
\end{equation}
\begin{equation} \textstyle
\small
        CL_{A_E}(x,k_{x}) =- \sum_{i=1}^{k_{x}}
        P_{A_E}(x,k_x,i)
        \ln(P_{A_E}(x,k_x,i))
\end{equation}
where $a^x_i$/$a^x_j$ represent the $i$/$j$th highest attribution score from the explanation set for $x$.

The final complexity score is an average of $CL_{A_E}(x,k_{x})$ across all $x \in D$:
\begin{equation}\textstyle
\small
CL_{E}={\sum_{x\in D}^{\left|D\right|}CL_{E}(x,k_x)}/{\left|D\right|}
\end{equation}
Notably, $k_{x}$ is calculated from the number of explanations produced by method $A_{\spanintex}$ for $x$, which varies based on the span interaction extraction method and is known only after generation.

\section{Experimental Setup}
\subsection{Datasets}
\label{Datasets}
We select the natural language inference dataset SNLI~\citep{bowman-etal-2015-large}, where instances consist of a premise, hypothesis, and a label $y\in \{entailment, neutral, contradiction\}$. 
%
Additionally, we select the fact checking dataset FEVER~\citep{thorne-etal-2018-fever,atanasova-etal-2020-generating}, where instances consist of a claim, evidence, and a label $y\in \{entailment, neutral, contradiction\}$.\footnote{
\url{https://huggingface.co/datasets/copenlu/fever_gold_evidence}} We generate input feature explanations by sampling 4,000 instances from each train, dev, and test set, due to the high computational cost, particularly for Shapley-based explanations \cite{atanasova-etal-2020-diagnostic}.
For Agreement with Human Annotations property, we use existing human explanation annotations for SNLI and FEVER (see \S\ref{sec:app:dataset-details}).
 
\subsection{Input Feature Explanation Methods}
\label{sec:PosthocExplainabilityTechniques}
To generate importance scores, we first select three common \tokenex\ techniques -- Shapley~\citep{lundberg2017unified}, Attention~\citep{deyoung-etal-2020-eraser}, and Integrated Gradients (IG, \citet{sundararajan2017axiomatic}). For \tokenintex, we employ Bivariate Shapley~\cite{masoomi2023explanations}, Attention~\citep{clark-etal-2019-bert}, and Layer-wise Attention Attribution~\citep{ye-etal-2021-connecting}. Notably, the \tokenintex\ techniques are the bivariate version of the techniques used for generating \tokenex; e.g. Layer-wise Attention Attribution uses IG.

Following~\citet{choudhury-etal-2023-explaining}, we apply the Louvain algorithm~\citep{blondel2008fast} for each of the three selected \tokenintex\ to generate importance scores for \spanintex\ methods, where the importance score of each span interaction is averaged over the importance scores of the token interactions within it. We will refer to Shapley, Attention, and IG as the explanation base types used for generating all types of input feature explanations for brevity. See \S\ref{sec:app:explainability_techniques} for more details.

\subsection{Models}
\label{Models}
We use the BERT-base-uncased model~\citep{devlin-etal-2019-bert} with 12 encoder layers, and the BART-base-uncased model~\citep{lewis-etal-2020-bart} with 6 encoder and 6 decoder layers, as common representatives of the encoder and the encoder-decoder Transformer architecture \cite{NIPS2017_3f5ee243}. Our choice is particularly influenced by the substantial computational requirements of the input feature explanations, especially pronounced for Shapley \cite{atanasova-etal-2020-diagnostic}. Additionally, our choice is guided by the need to directly access the models' internals for generating IG and Attention-based explanations. Furthermore, while our framework currently utilizes the said models, it is designed to be easily adaptable to other models or newly developed explainability techniques, provided that there are more robust computational resources available. 

We fine-tune the base models on SNLI and FEVER and use them to generate explanations. Their performance on the test sets is shown in \S\ref{sec:app:base-model}. For the Simulation property, we follow existing work \cite{fernandes2022learning,pruthi-etal-2022-evaluating} and train simulator agent models (\S\ref{sec:simulatability:method}) with the same architectures as the base ones. Following \citet{fernandes2022learning}, we split the test set into train/dev/test for the training of the agent model.

\section{Results and Discussion}

\begin{figure*}[t]
\centering
\subfigure[SNLI dataset, BERT model.]{\label{fig:subfig:a}
\includegraphics[width=0.49\linewidth]{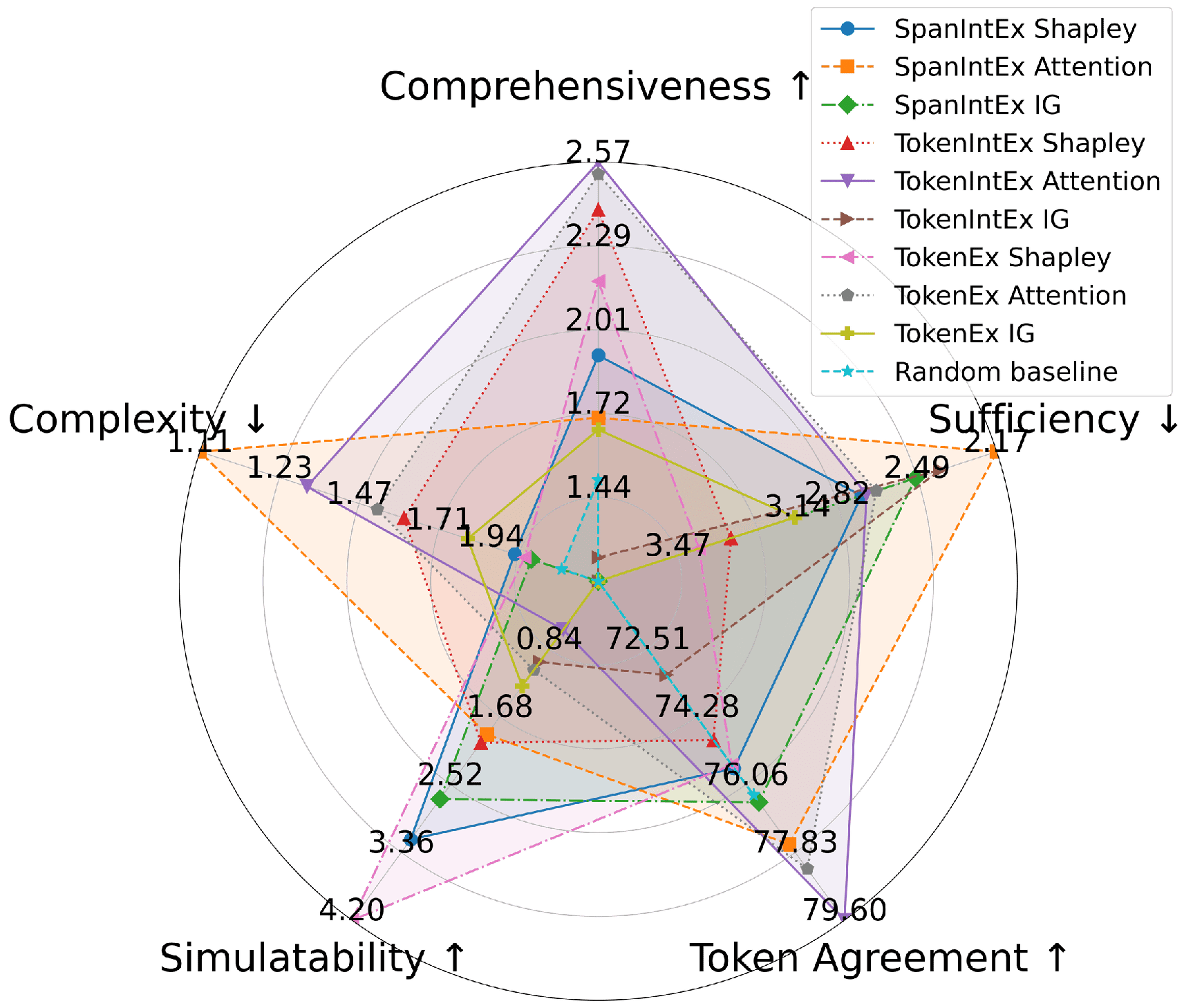}}
\subfigure[SNLI dataset, BART model.]{\label{fig:subfig:b}
\includegraphics[width=0.49\linewidth]{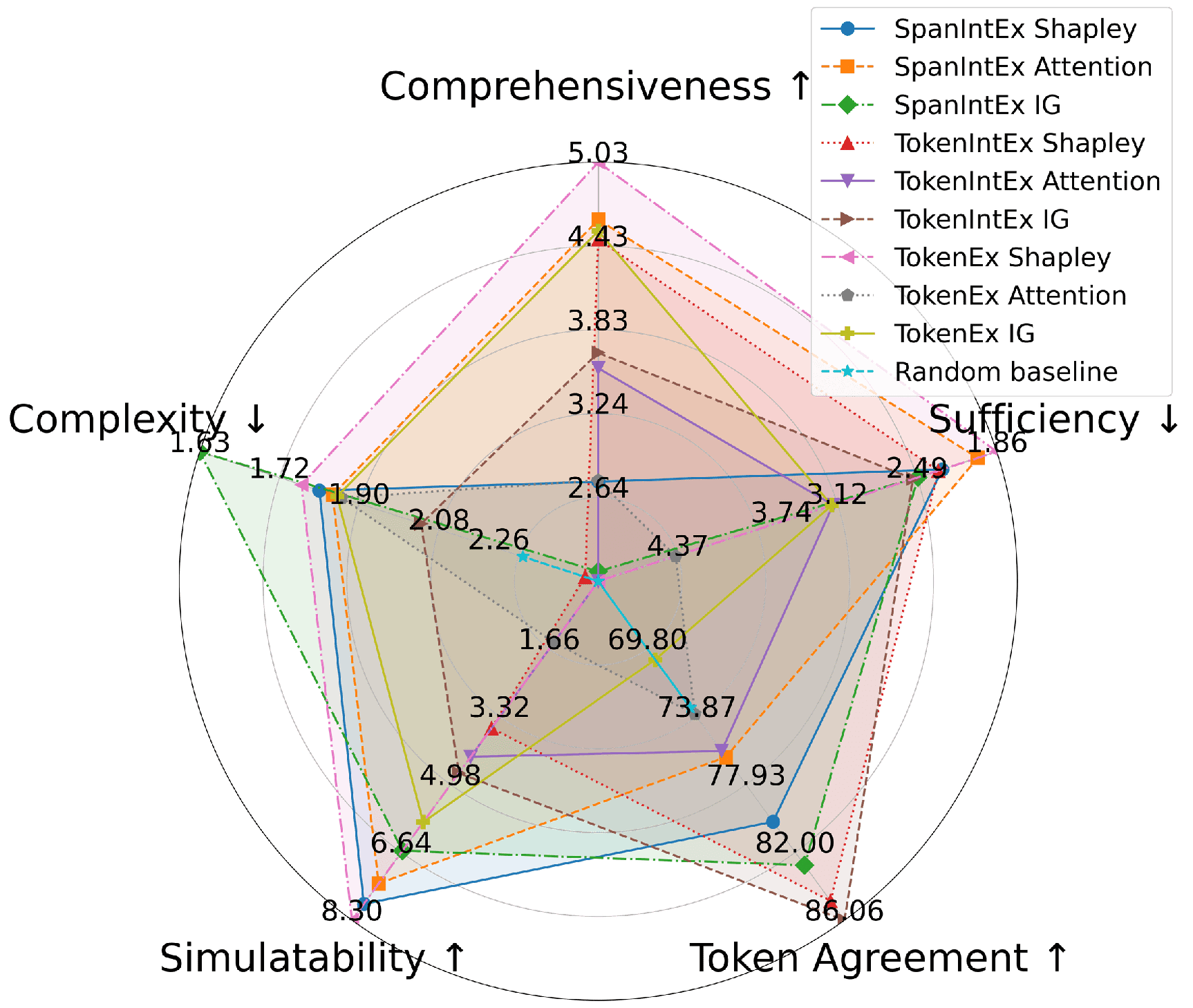}}
\subfigure[FEVER dataset, BERT model.]{\label{fig:subfig:c}
\includegraphics[width=0.49\linewidth]{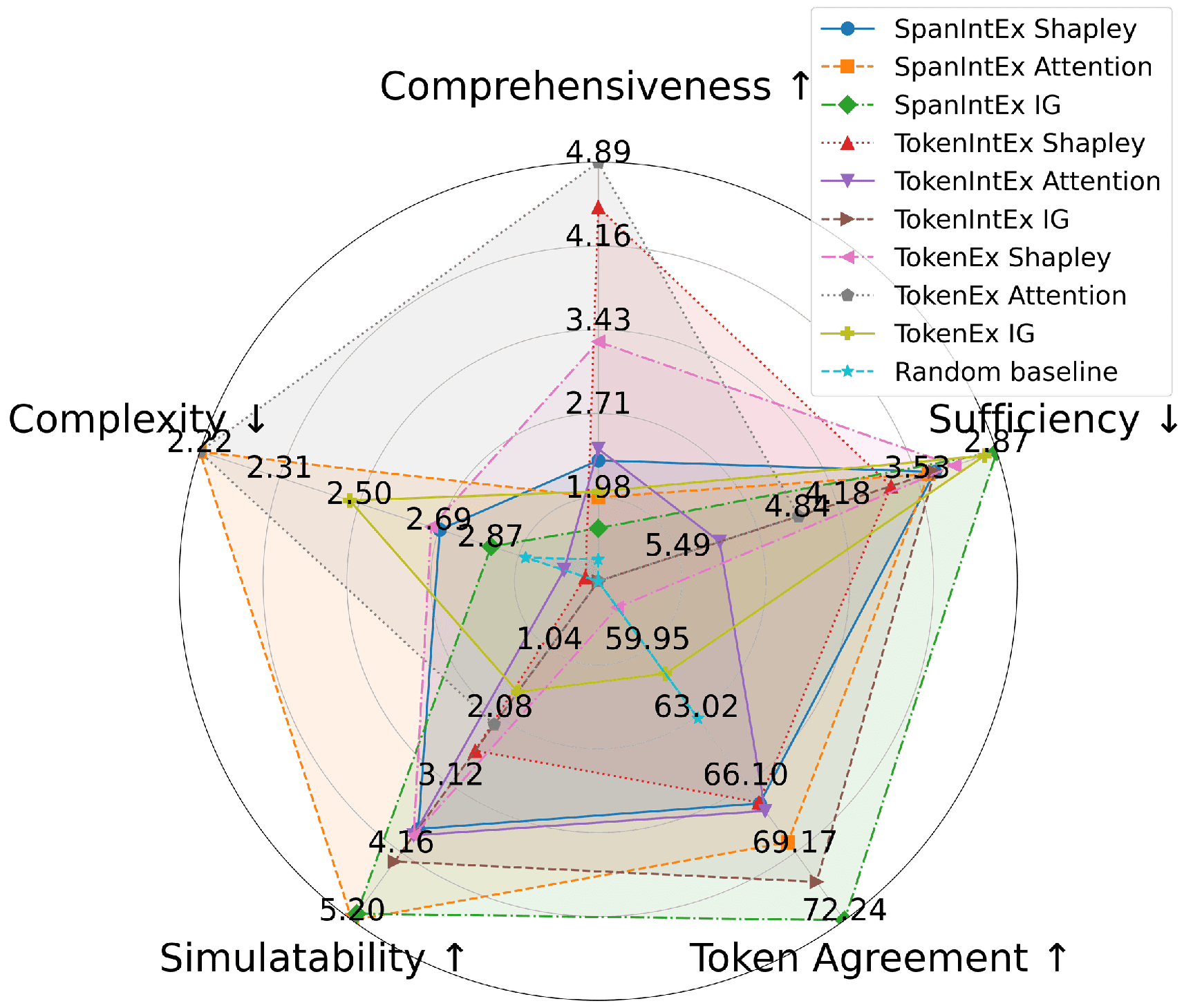}}
\subfigure[FEVER dataset, BART model.]{\label{fig:subfig:d}
\includegraphics[width=0.49\linewidth]{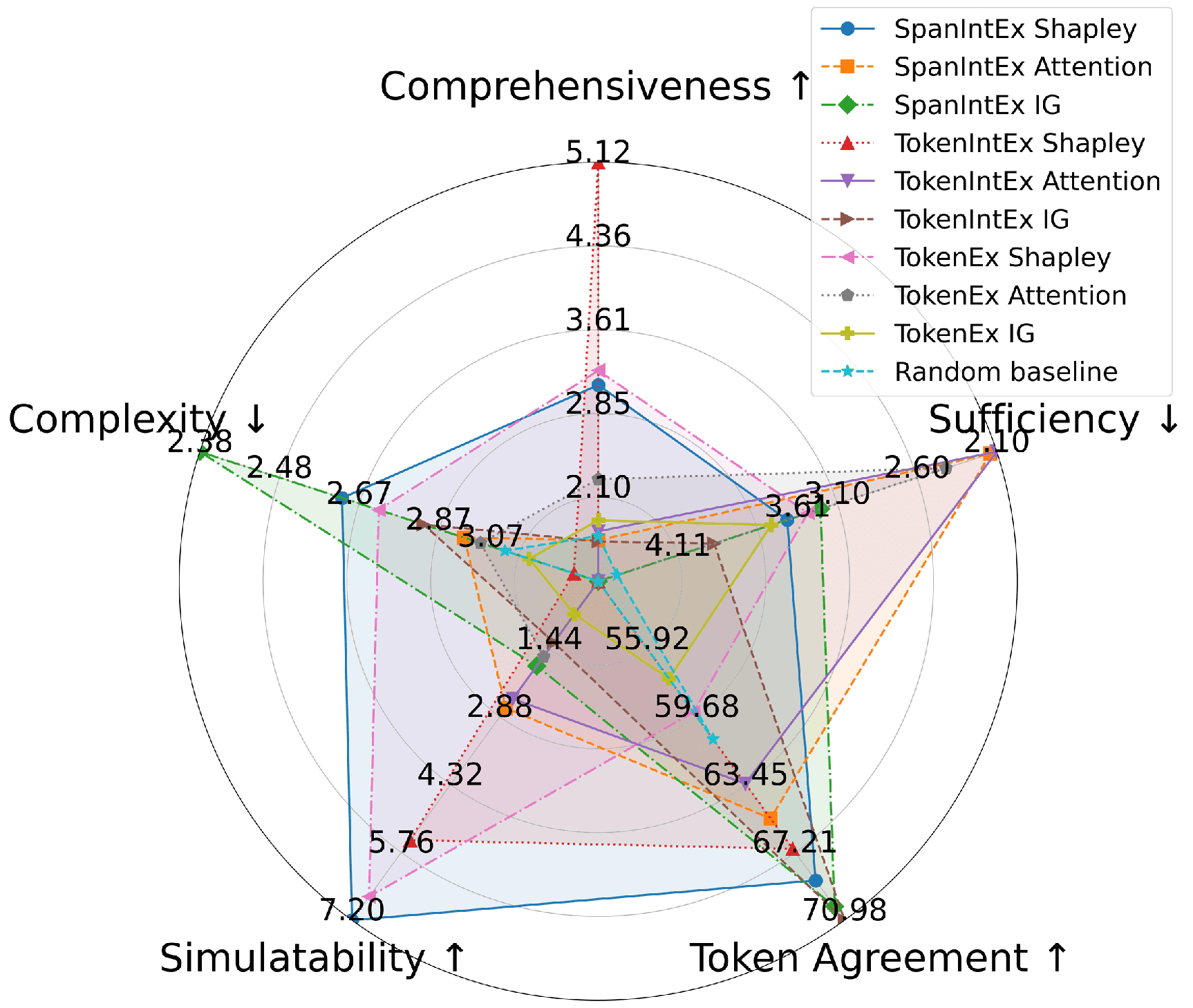}}
\caption{Unified evaluation framework (\S\ref{sec:methodology:overall}) results for all feature attribution methods (\S\ref{sec:PosthocExplainabilityTechniques}).}
\label{fig:mainfigure}
\scriptsize
\end{figure*}

We now present the results of our unified evaluation framework (\S\ref{sec:methodology:overall}) illustrated in Fig. \ref{fig:mainfigure}. They include explanation methods of types \spanintex, \tokenintex, and \tokenex\ (\S\ref{sec:PosthocExplainabilityTechniques}), two models (\S\ref{Models}), two datasets (\S\ref{Datasets}), and three base explanation techniques per explanation type (\S\ref{sec:PosthocExplainabilityTechniques}). For Simulatability, we select the results of $I_{sym}$, as this form avoids repeating the input text and increasing the input size substantially. For Agreement with Human Annotations, we select the Token-level results as they are present for all explanation types. \S\ref{sec:app:detailed} lists detailed results per property.

\subsection{\textbf{Faithfulness}}
\label{result:faithfulness}
\textbf{Unified Comprehensiveness}. Across both datasets and models, \textbf{\tokenex\ and \tokenintex\ are identified as the most comprehensive explanation types}, achieving the highest scores in 7/12 and 5/12 cases, respectively. \spanintex, designed to enhance the semantic coherence of interactive explanations by including additional context, often incorporates tokens that do not directly contribute to the model's prediction, thus explaining its lower comprehensiveness scores. Compared to the random baseline, \tokenex\ and \spanintex\ always outperform it, while \tokenintex\ mostly underperform it when based on IG. Across the base explanation techniques, \tokenex\ performs best when based on Attention for BERT and on Shapley for BART, indicating that \textbf{different base explanation techniques can perform better for different architectures}. Both \tokenintex\ and \spanintex\ 
show optimal performance when based on Shapley and Attention. Overall, the results indicate a \textit{stronger performance of Attention and Shapley over IG} across all explanation types. 

\textbf{Unified Sufficiency.} 
\textit{\spanintex\ ranks as the most sufficient explanation type in 7/12 cases}, surpassing \tokenex, which performs well in only 3/12 cases.  While contrary to \spanintex\ Comprehensiveness performance, we attribute this to the semantic coherence of the extracted top spans, which provide more meaningful information. 
Note that while Sufficiency is highly desirable, Comprehensiveness is not required in all downstream applications as end-users prefer simpler, more general explanations with fewer causes \cite{thagard1989explanatory}.
Unlike \tokenex, which consistently outperforms the random baseline, \tokenintex\ and \spanintex\ struggle to outperform it on FEVER, likely due to the longer inputing challenges for the explanations to accurately unveil the model’s internal processes. 
The results from different base explanation techniques show no clear trends, indicating a \textbf{significant variability stemming from the specific dataset and model architecture}.

\subsection{\textbf{Agreement with Human Annotation}}
\label{result:agreement}
\spanintex\ and \tokenintex\ show higher agreement scores with human annotations than \tokenex. Similarly, \spanintex\ consistently achieve higher agreement with human interaction-level annotators, especially when based on Attention scores (see 
\S\ref{sec:app:detailed}). This indicates that \textbf{\spanintex\ are more plausible to humans due to their enhanced semantic coherence}. In contrast, \tokenex\ often score lower than the random baseline. Moreover, considering \spanintex's lower performance in Comprehensiveness, there emerges \textbf{a distinct trade-off between Comprehensiveness and Agreement with Human Annotations}. Across base explanation techniques, IG performs best for FEVER; IG and Attention -- for SNLI. 
In addition, we find that for the interaction-level agreement, \tokenintex\ and \spanintex\ perform worst when based on Shapley. The lower agreement results for Shapley compared to its better results on Comprehensiveness again indicate an existing trade-off between the two properties.

\subsection{\textbf{Simulatability}}
\label{result:simulatability}
Our results show that \spanintex\ achieve the highest simulatability in 9/12 cases, helping the agent model accurately reproduce the original model's prediction. This again underscores the \textbf{critical role of contextual information and enhanced semantic coherence provided by \spanintex}. Notably, providing \spanintex\ to agents improves their ability to simulate the original model by up to 7.9 F1 points compared to without explanations. Among base explanation techniques, IG consistently performs best for \spanintex; other techniques do not exhibit a clear trend.

\subsection{\textbf{Complexity}}
\label{result:comlexity}
\spanintex\ and \tokenex\ generally achieve similar complexity, which consistently remains lower than those of \tokenintex. This suggests that \textbf{\tokenex\ and \spanintex\ generate more distinctive attribution scores, potentially making them easier for humans to understand}. 
Regarding the base explanation techniques, Attention consistently yields the best complexity scores for BERT across all explanation types. There is no clear trend for BART. Additionally, \tokenintex\ frequently underperform the random baseline, highlighting its complexity (see \S\ref{sec:app:detailed}).

\subsection{Overall}
In summary, we find that while \tokenex\ and \tokenintex\ generally provide more comprehensive insights, \spanintex\ performs better in Sufficiency due to its enhanced semantic coherence (\S\ref{result:faithfulness}). 
This calls for \textit{better methods for generating \spanintex\ that are both comprehensive and sufficient}. 
Additionally, there is a trade-off between Comprehensiveness and Agreement with Human Annotations (\S\ref{result:agreement}), suggesting that the most faithful explanations might be less plausible to end users. This highlights the \textit{need for advanced methods to boost both the Comprehensiveness and plausibility} of \spanintex\, possibly leveraging the advantage of \tokenex. Furthermore, \spanintex\ significantly improves simulatability by allowing agents to accurately replicate model decisions (\S\ref{result:simulatability}), which is crucial in practice. Finally, the complexity analysis (\S\ref{result:comlexity}) shows that \spanintex\ and \tokenex\ are potentially easier to comprehend than \tokenintex\ when considering the importance score distribution. 

Overall, our results highlight the differences between the different types of input feature explanations, with \spanintex\ outperforming others on most measures. As no one type performs best on all properties, we \textit{call for the development of combined methods that can leverage the strength of the different explanation types and potentially lead to an overall improvement of the explanation utility}.

\section{Related Work}

\textbf{Input Feature Explanations.} Considerable research exists on extracting explanations for input data.
Methods like perturbation-based attribution (e.g., Shapley~\citep{lundberg2017unified}), attention-based methods (e.g., Attention~\citep{jain-wallace-2019-attention,serrano-smith-2019-attention}), and gradient-based methods (e.g., Integrated Gradients~\citep{sundararajan2017axiomatic,serrano-smith-2019-attention}) are prevalent for highlighting individual tokens~\citep{atanasova-etal-2020-diagnostic}. As individual tokens might be insufficient to explain the model, many attribution methods have been extended to bivariate forms~\citep{masoomi2023explanations,janizek2021explaining,sundararajan2017axiomatic,ye-etal-2021-connecting} to capture input token interactions. More recent work has explored how interactions between groups of tokens collectively contribute to model reasoning~\citep{choudhury-etal-2023-explaining,chen-etal-2021-explaining}. Unlike other work where token groups might consist of tokens from arbitrary positions, \citet{choudhury-etal-2023-explaining} explicitly capture span interactions, enhancing the comprehensiveness of explanations by containing the entire spans. 

\textbf{Automated Explanation Evaluation.} For evaluating \tokenex, \citet{deyoung-etal-2020-eraser,atanasova-etal-2020-diagnostic} propose metrics to measure how faithful explanations are to the model's inner reasoning. They also assess the plausibility of explanations to humans by measuring the agreement of \tokenintex\ with human annotations. To assess the utility of explanations to humans, \citet{pruthi-etal-2022-evaluating} propose to use an agent model as a proxy for humans and evaluate whether explanations aid in model simulatability. Complexity \citet{10.5555/3491440.3491857} measures the distribution of attribution scores of \tokenex\ and assesses whether the key tokens in token explanations are easily comprehensible to humans. 
To evaluate \tokenintex\, most works adopt the faithfulness or axiomatic/theoretical path ~\cite{tsang2020does,sundararajan2020shapley,janizek2021explaining}.
Current work on evaluating \spanintex\ has primarily focused on faithfulness \cite{choudhury-etal-2023-explaining}. However, since \spanintex, \tokenintex, and \tokenex\ contain varying amounts of tokens, which, e.g., affects the faithfulness test, this makes direct comparisons between different explanation types using existing metrics challenging. To our knowledge, no prior paper has involved all types of input feature explanations within a unified evaluation framework.

\section{Conclusion}
We introduced a unified evaluation framework for input feature attribution analysis to guide the principled selection of the most suitable explainability technique in practical applications. Our analysis outlines the diverse strengths and trade-offs among \tokenex, \tokenintex, and \spanintex. Our findings particularly underscore \spanintex's superior performance in Sufficiency, agreement with human inductive biases, its enhancement of Simulatability, and Complexity, compared to \tokenex\ and \tokenintex. Future efforts should focus on developing combined methods that enhance all explanation properties.

\section*{Limitations}
Our work introduces a unified framework to evaluate input feature explanations across four key properties. We generated three types of explanations using three attribution methods on two Transformer models (BERT-base and BART-base) for two NLU tasks (NLI and fact-checking). Thereby, we can assess and compare the properties of each explanation type. Due to computational resource limitations, we did not include larger decoder-only models in our evaluation. Future research could \textit{explore other models} to provide additional insights. 

We note that our work considered the FEVER and SNLI datasets as they are the only available datasets with annotations of human interactive explanations, required for the Agreement with Human Annotations property. In future work, given the availability of other datasets, examining the properties of different explanations in various \textit{tasks beyond NLI and fact checking} would be valuable, especially for simpler tasks that consist of only one input part or more complex tasks that consist of more than two parts with possible relationships between them. Additionally, tasks with longer textual inputs, which are known to pose greater challenges for current explainability techniques~\citep{atanasova-etal-2020-diagnostic}, could also be analyzed. 

Furthermore, while we consider four widely used explanation properties in this automatic evaluation framework, future works should consider verifying, potentially with supplementary human studies, that the properties are well aligned with the downstream utility of the explanations in different application tasks~\citep{miller2019explanation}. We note that manual evaluation, while valuable, is time-consuming and costly. Automated evaluation, with our proposed framework, allows for quicker insights, helping prioritize methods that may benefit from human-centered evaluation. Additionally, the properties we evaluate demonstrate why human evaluation is not necessarily required at this stage. Faithfulness measures whether explanations reflect the model’s internal reasoning, a task humans cannot assess (see Faithfulness evaluation guidelines in ~\citet{jacovi2020towards}). Explanations that fail this test should not be considered for further human evaluation as they can be harmful, e.g. by hiding a model's flaws and biases. Agreement with Human Annotation already captures alignment with human reasoning, ensuring explanations are plausible. Automated Simulatability correlates strongly with human studies, providing a reliable proxy for replicating the model’s behaviour without the need for expensive human experiments~\citep{pruthi-etal-2022-evaluating}. Finally, automated methods ensure consistency and objectivity, while human annotations can introduce subjectivity and variability. Studies have even shown conflicting results from human evaluations~\citep{poursabzi2021manipulating,ribeiro2016should}. Automated evaluation provides an objective, reproducible baseline, which can later be supplemented by human evaluations where needed.

We have also employed three base representative explainability methods for each of the three types of input feature explanations. However, \textit{more existing and newly emerged base explainability methods} could be explored in future work. Additionally, our study focuses solely on post-hoc explainability techniques, while other supervised feature extraction methods could also be investigated~\citep{yu2021understanding,liu2024mmi}.  These methods typically treat human-annotated important fragments within the input as gold causal features, akin to our Agreement with Human Annotation measurement. The broader set of properties introduced in our framework could be leveraged to evaluate such explanations more comprehensively. Apart from that, future work could adapt and extend our framework to other forms of explanations such as free-text explanations produced by self-rationalization models ~\citep{liu2022fr,liu2023decoupled,liu2023mgr,liu2024d,liu2024enhancing}. All said potential future studies are well facilitated by the efficient automated evaluation proposed with our framework.

Our findings indicate that span interactive explanations (\spanintex) have a notable advantage over other explanation types in terms of Agreement with Human Annotation, Simulatability, and Complexity, suggesting they are easier for humans to understand. This insight could inspire future work to leverage \spanintex\ as the input feature explanation in HCI models. However, \spanintex\ shows low comprehensiveness in faithfulness evaluations. The Louvain algorithm, used for \spanintex\ generation, may limit its comprehensiveness despite using different attribution methods for \tokenintex. Future work should explore better methods for capturing span interactions and possibly combine \spanintex\ and \tokenex\ for higher faithfulness, as \tokenex\ demonstrates a stable advantage in comprehensiveness.

Another core finding is that the attribution method significantly affects most diagnostic properties of all explanation types, such as sufficiency. No single attribution method consistently excels across all properties, highlighting the need for continuous evaluation and improvement in attribution methods, particularly for \spanintex.

To ensure a fair comparison, our unified evaluation framework currently considers only the token count differences among various input feature explanations, with interactive explanations flattened. Future work could involve a human-in-the-loop approach to account for the effects of interactive explanations beyond just token count differences. For example, a display system could visually present highlighted tokens and interactions to gauge human preferences. Our work provides a starting point for comparing input feature explanations from an automated evaluation perspective, and future research could explore additional factors, such as psychological elements and visual aspects, from a human perspective, which would benefit more non-expert users.

Another limitation of this work is that we focus solely on the automatic evaluation of input feature explanations without examining the potential biases they may exhibit. For instance, these explanations might favor certain words or phrases from the input sequence, for example, sometimes emphasizing some prepositions that might bear less meaning, raising questions about whether such biases stem from the models themselves or from the explainability techniques used. Also, it is worth checking, especially in sensitive domains such as healthcare and law, how contradictory explanations for different model decisions differed, which might sway the decision-making. 

\section*{Acknowledgements}
$\begin{array}{l}\includegraphics[width=1cm]{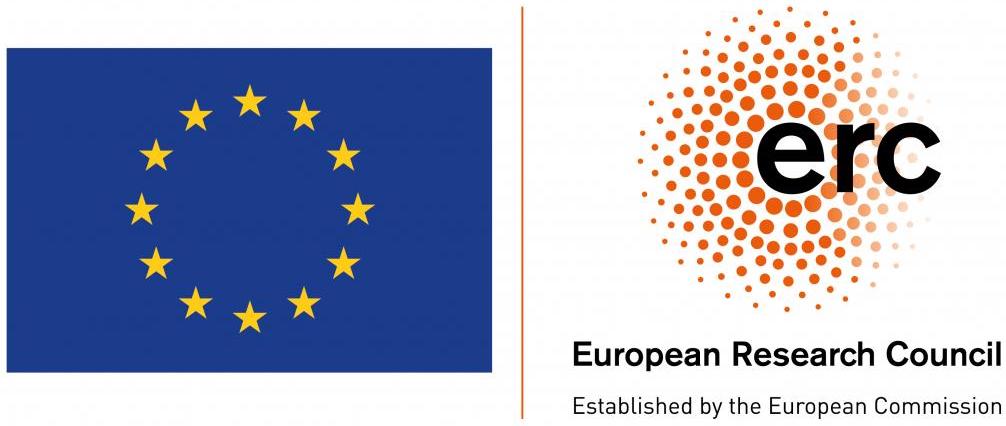} \end{array}$ 
This research was co-funded by the European Union (ERC, ExplainYourself, 101077481), by the Pioneer Centre for AI, DNRF grant number P1, as well as by The Villum Synergy Programme (grant number 40543). Views and opinions expressed are however those of the author(s) only and do not necessarily reflect those of the European Union or the European Research Council. Neither the European Union nor the granting authority can be held responsible for them.

\bibliography{anthology,custom}
\bibliographystyle{acl_natbib}
\clearpage
\appendix

\section{Unified Faithfulness Evaluation: Explanation Masking Details}
\label{sec:app:faithfulness-mask-details}
As discussed in Section \S\ref{sec:faithfulness:method}, we introduce the dynamic threshold $\theta_{x,k}$ to ensure an identical number of tokens from different types of top input feature explanations for the same instance in Unified Faithfulness evaluation.

For the Unified Comprehensiveness evaluation, we conduct a similar process for all three explanation types separately. Using \tokenex\ as an example, we first calculate the maximum number of top \tokenex\ for disturbing each instance $x$ as $k_{\tokenex}(x)$ following Eq. \ref{eq:threshold_2}. To omit the \tokenex\ tokens from the original input, we replace them with [MASK] tokens, while keeping the rest unchanged. The specific [MASK] token used depends on the model architecture. We then gradually increase the number of \tokenex\ tokens masked out until it reaches  $k_{\tokenex}(x)$ and record the corresponding changes in model predictions. The average prediction change across the dynamic threshold and all instances gives the Unified Comprehensiveness score for \tokenex. For \tokenintex\ and \spanintex, the only difference is that we mask out token pairs for \tokenintex\ and span pairs for \spanintex, with the maximum explanations masked out as $k_{\tokenintex}(x)$ and  $k_{\spanintex}(x)$ for each instance $x$, calculated using Eq. \ref{eq:threshold_1} and Eq. \ref{eq:threshold_3} respectively.

For the Unified Sufficiency evaluation, we conduct experiments for the three explanation types separately. Unlike Unified Comprehensiveness, we retain only the tokens/token pairs/span pairs for the input while masking out all other tokens with [MASK] tokens for each instance, depending on the model architecture used. We first calculate the maximum number of top explanations involved in disturbance for each explanation type for instance $x$ using Eq. \ref{eq:threshold_1}, Eq. \ref{eq:threshold_2}, and Eq. \ref{eq:threshold_3}. Then, we keep the token/token pairs/span pairs in the model input by masking out all other tokens, starting with one explanation and adding one more explanation for each subsequent disturbance until the total number of explanations reaches $k_{\tokenex}(x)$, $k_{\tokenintex}(x)$, or $k_{\spanintex}(x)$. Meanwhile, we record the model predictions for each disturbance. The Unified Sufficiency score for each explanation type is then calculated by averaging the prediction changes across the dynamic threshold for that explanation type, considering all instances.

\section{Detailed Explanation Insertion Method}
\label{sec:app:detailed-form}
To enable a fair comparison among different input feature explanations in terms of simulatability (\S\ref{sec:simulatability:method}), we applied consistent insertion formats to combine the explanations with the original input for training the agent models. This design aims to minimize noise from insertion format differences. We tested two ways, each applicable to all types of input feature explanations, to construct input sequences with inserted explanations of type $E$. These input sequences are denoted $x_E$ in \S\ref{sec:simulatability:method}, omitting specific insertion format details for brevity.

For Symbol-Insertion \(I_{Symbol}\), we preserve the original input sequence but insert special symbols \(<\) and \(>\) to quote the tokens (for \tokenex\ and \tokenintex) or spans (for \spanintex) within the input. Additionally, for \tokenex, we append a ranking mark after each quoted token based on their attribution scores, ranked in descending order. For \tokenex\ and \spanintex, each quoted token/span is also assigned a ranking mark indicating the rank of their respective interactions by attribution score, ensuring tokens/spans from the same interaction share the same mark. This method allows us to generate input sequences combined with different input feature explanations in a consistent symbol insertion format.

For Text-Insertion \(I_{Text}\), we append tokens, token tuples, or span tuples to the end of the original input sequence for each explanation type. They are added in the order ranked by descending attribution score. Specifically, for \tokenex, tokens from different \tokenex explanations are separated by semicolons. For \tokenintex\ and \spanintex, tokens/spans within each interaction are connected by a comma, and different interactions are separated by semicolons. This approach constructs input sequences combined with each type of input feature explanation in a consistent text insertion format.

\section{Agreement Dataset Details}
\label{sec:app:dataset-details}

To assess how different types of input feature explanations overlap with human annotations, we collected golden explanations of various types from e-SNLI and SpanEx for instances within SNLI and FEVER, respectively. Detailed information about the annotated explanation types and the number of instances with labeled explanations for these datasets is shown in Table \ref{exper:dataset-details}. For the SNLI dataset, e-SNLI provides \tokenex\ explanations, while \textit{SpanEx-SNLI} includes \spanintex\ explanations and token-level interactions (\tokenintex\ explanations). We selected 3,865 overlapping instances and evaluated the human agreement score for different types of input feature explanations. For FEVER, \textit{SpanEx-FEVER} includes \spanintex\ and token-level interactions (\tokenintex\ explanations). Since no \tokenex\ explanations are provided, we extracted tokens from the golden \tokenintex\ explanations in \textit{SpanEx-FEVER} as an approximation. These selected instances are also used when evaluating other properties of input feature explanations.

\begin{table}
\centering
\small
\begin{tabular}{lcccc}
\toprule
\textbf{$D$} & $E$ & \multicolumn{1}{l}{\textbf{Size}}                                           \\  \midrule
SNLI        & -   & \begin{tabular}[c]{@{}c@{}}549367 Train\\ 9842 Dev\\ 9824 Test\end{tabular} \\
e-SNLI                & \tokenex\                                                         & \begin{tabular}[c]{@{}c@{}}549367 Train\\ 9842 Dev\\ 9824 Test\end{tabular} \\
\textit{SpanEx-SNLI}  & \begin{tabular}[c]{@{}c@{}}\spanintex\ \\ \tokenintex \end{tabular}  & 3865 Test                                                                   \\  \midrule
FEVER        & -   & \begin{tabular}[c]{@{}c@{}}145449 Train\\ 9999 Dev\\ 9999 Test\end{tabular} \\
\textit{SpanEx-FEVER} & \begin{tabular}[c]{@{}c@{}}\spanintex \\ \tokenintex \end{tabular} & 3206 Test                                                                   \\ \bottomrule
\end{tabular}%
\caption{Overview of datasets SNLI~\citep{bowman-etal-2015-large}, FEVER~\citep{thorne-etal-2018-fever}, \textit{SpanEx}~\citep{choudhury-etal-2023-explaining} and e-SNLI~\citep{camburu2018snli}. \textit{SpanEx} contains instances from SNLI and FEVER, annotated with \spanintex\ explanations including token-level explanations (\tokenintex\ explanations). e-SNLI contains instances from SNLI dataset, annotated with \tokenex\ explanations.}
\label{exper:dataset-details}
\end{table}

\section{Base Model Performance.}
\label{sec:app:base-model}
As shown in Table \ref{exper:base-model-performance}, we report the performance of fine-tuned BERT-base and BART-base models on SNLI and FEVER, respectively. These models, fine-tuned for their specific tasks, are used to generate various input feature explanations through different explainability techniques. Importantly, these are the original models that the agent models, as described in \S\ref{sec:simulatability:method}, learn to simulate.

\begin{table}
\centering
\small
\begin{tabular}{lcccc}
\toprule

\multirow{2}{*}{\textbf{$Model$}} & \multicolumn{2}{c}{\textbf{F1 score}}              \\
                                  & \multicolumn{1}{l}{Dev} & \multicolumn{1}{l}{Test} \\ \midrule
BERT-SNLI                         & 87.21                   & 88.43           \\
BART-SNLI                         & 86.81                   & 85.40                    \\ \midrule
BERT-FEVER                        & 86.21          & 89.49           \\
BART-FEVER                        & 85.19                   & 84.88                    \\ \bottomrule
\end{tabular}%

\caption{The performance of our BERT-base and BART-base models fine-tuned on SNLI and FEVER, respectively, regarding F1 score(\%).}
\label{exper:base-model-performance}
\end{table}

\section{Explainability Techniques}
\label{sec:app:explainability_techniques}

In this section, we detail the explainability techniques employed to generate various types of input feature explanations. As outlined in \S\ref{sec:PosthocExplainabilityTechniques}, we categorize these techniques based on the method used for generating \tokenex, while \tokenintex\ explanations stem from their bivariate variants, forming the basis for \spanintex\ explanations.

As denoted in Section \S\ref{preliminaries}, $x_i$ represents the $i$th token with instance $x$. To better illustrate the explainability techniques below, we use $F$ as the set of all tokens within this instance and $S$ as the subset of $F$. All explanations are obtained using model $M$, which is omitted in the following notions for brevity. We use $A_\tokenex(x_i)$ to denote the attribution score generated by explainability technique $A$ for the $i$th token $x_i$, $A_\tokenintex(x_i,x_j)$ as the attribution score for token interaction $(x_i,x_j)$,  $A_\tokenintex(x_i \mid x_j)$ as the importance score of token $x_i$ conditioned on $x_j$ is present when the directed importance between tokens within $(x_i,x_j)$ is considered in some attribution techniques,
$A_\spanintex(span_i^0,span_i^1)$ as the attribution score for corresponding span interaction, where $span_i^0=(x_s,...,x_{s+l_1})$ is a span from part1 and $span_i^1=(x_t,...,x_{t+l_2})$ is a span from part2 of the input. Note that in Section \S\ref{preliminaries}, we use $a^x_k$ to denote the importance score of the $k$th most important explanation of instance $x$; here, we only focus on the attribution scores of explanations without ranking them. 

\textbf{Shapley}. For \tokenex,  we employ the SHAP method to assign importance scores to each token within the input by removing each token separately and computing its removal effect on the model prediction with different subsets of other tokens presented to the model, following \citet{lundberg2017unified}. 
\begin{equation}
\small
\begin{split}
Shap_\tokenex(x_i) &= \sum_{S \subseteq F \setminus \{x_i\}} 
\frac{|S|!(|F| - |S| - 1)!}{|F|!} \\
&\quad [f(S \cup \{x_i\}) - f(S)]
\end{split}
\label{eq:univariate_shapley}
\end{equation}
$Shap_\tokenex(x_i)$ denotes the importance score of token $x_i$.
As the calculation of $Shap_\tokenex(x_i)$ is computationally expensive, we utilize Kernel SHAP to approximate these Shapley values. 

For \tokenintex, we first apply Bivariate Shapley ~\citep{masoomi2023explanations} to assess the mutual importance scores between two tokens, which are from different parts of the input, within a token interaction, and then average these two mutual importance scores as importance score of this token interaction. Specifically, to compute the importance score of a token $ x_i$ conditioned on the presence of token $ x_j$, the sets of tokens $S$ considered are limited to those containing token $x_j$, while the impact of other sets of tokens influences the importance of $x_i$ is ignored in this case. Thus, $Shap_\tokenintex(x_i \mid x_j)$ can be calculated by:


\begin{equation}
\scriptsize
\begin{split}
Shap_\tokenintex(x_i \mid x_j) &= \sum_{x_j \in S \subseteq F \setminus \{x_i\}} 
\frac{|S|!(|F| - |S| - 1)!}{|F|!} \\
&\quad [f(S \cup \{x_i\}) - f(S)]
\end{split}
\label{eq:bivariate_shapley_directed}
\end{equation}
The importance score $Shap_\tokenintex(x_i,x_j)$ for token interaction $(x_i,x_j)$ are calculated by averaging $Shap_\tokenintex(x_i\mid x_j)$ and $Shap_\tokenintex(x_j\mid x_i)$. We also use Kernel Shapley to approximate the calculation of Bivariate Shapley value.

For \spanintex, we first apply the Louvain Community Detection algorithm ~\citep{blondel2008fast} to extract span interactions and then average the importance scores of token interactions comprised in each span interaction as its importance score, following \citet{choudhury-etal-2023-explaining}.

To extract span interactions, we first construct a directed bipartite graph for instance $x$, by taking each token $x_i$ from the input as node $i$ and the mutual importance scores between each two tokens from different parts obtained above as the weights of directed edges connecting them. Louvain Community Detection algorithm is then applied to search for communities of nodes with dense intra-cluster and sparse inter-cluster relationships. With each community of nodes(tokens) $S_p$ obtained, we can get one span interaction $(span_p^0,span_p^1)$, where the two spans consist of neighboring tokens from the part1 subset of this community $S_p^0$, and the part2 subset of it $S_p^1$ respectively. 

Then we calculate the importance score of this span interaction by averaging the importance scores of all token interactions it comprises.
\begin{equation}
\scriptsize
\begin{aligned}
Shap_\spanintex(span_p^0,span_p^1) &= \\
&\quad \frac{\sum\limits_{i,j}^{\begin{array}{l}
    \substack{x_i \in S_p^0  \\ x_j \in S_p^1}
\end{array}} Shap_\tokenintex(x_i, x_j)}{|S_p^0 \cup S_p^1|}
\end{aligned}
\label{eq:span_extraction_method}
\end{equation}

Note that in the following, no matter which explainability techniques to assign importance score to \tokenintex, we apply the same method as stated above to extract span interactions and compute their importance scores, $A_\spanintex(span_p^0,span_p^1)$, based on corresponding token interaction importance score, $A_\tokenintex(x_i,x_j)$.

\textbf{Attention}. For each token within the input sequence, we use the self-attention weights between this token and the first token as an indicator of its importance score \citep{jain-wallace-2019-attention}. We follow \citet{choudhury-etal-2023-explaining} to select the most important attention head in the last layer of the model to obtain these attention weights. For each possible token interaction, we use the method by \citet{clark-etal-2019-bert} to extract and average the attention weights between token pairs from different parts of the input to derive their importance scores, also from the most important head of the last layer. To obtain span interactions and assign them importance scores, we apply the same method to these token interaction scores as described above.

\textbf{Integrated Gradients}. To calculate the importance score for each token in the input sequence, we integrate the gradients of the model's output with respect to each token embedding, following \citet{sundararajan2017axiomatic}. For generating the importance scores of token interactions, we use Layer-wise Attention Attribution \citep{ye-etal-2021-connecting}, which attributes attention links between pairs of tokens within attention maps with a mechanism similar to Integrated Gradients. These attribution maps are created for each model layer and then aggregated across layers to form a final attribution map. The importance score for each token interaction is calculated as the average value from this final attribution map between the involved tokens. For span interactions, we generate and assign importance scores using the same approach based on the importance scores of the token interactions.

\section{Detailed Experiment Results}
\label{sec:app:detailed}
\subsection{Faithfulness}

\begin{figure*}[t]
\centering
\subfigure[Comprehensiveness on SNLI dataset, the higher the better.]{\label{fig:subfig-comp-snli-bert:a}
\includegraphics[width=0.49\linewidth]{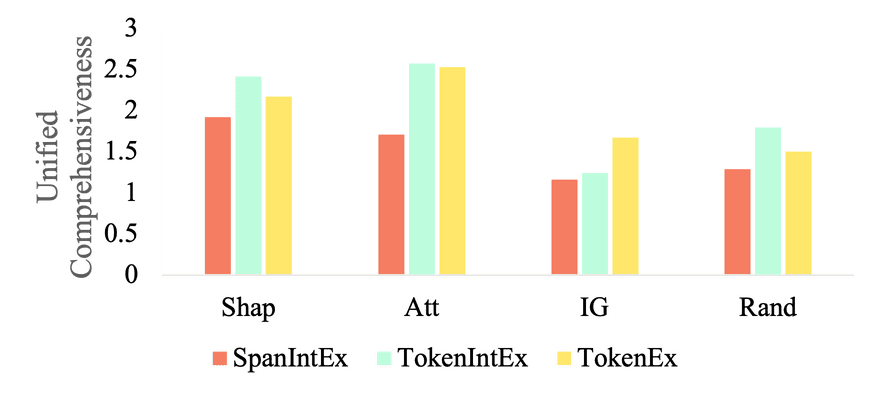}}
\subfigure[Comprehensiveness on FEVER dataset, the higher the better.]{\label{fig:subfig-comp-fever-bert:b}
\includegraphics[width=0.49\linewidth]{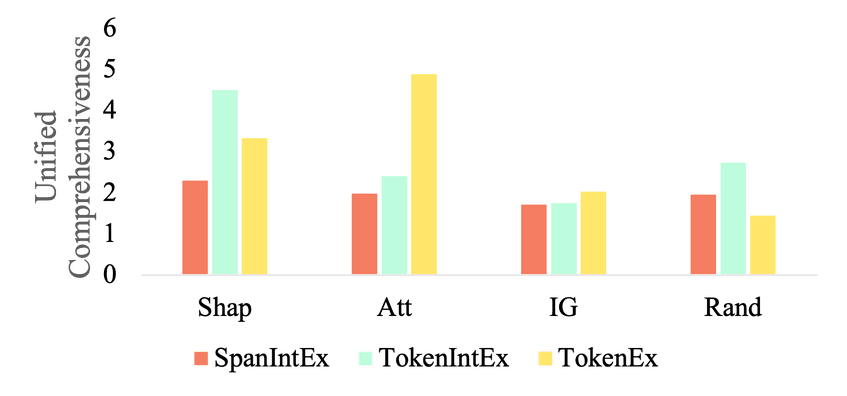}}
\subfigure[Sufficiency on SNLI dataset, the lower the better.]{\label{fig:subfig-suff-snli-bert:c}
\includegraphics[width=0.49\linewidth]{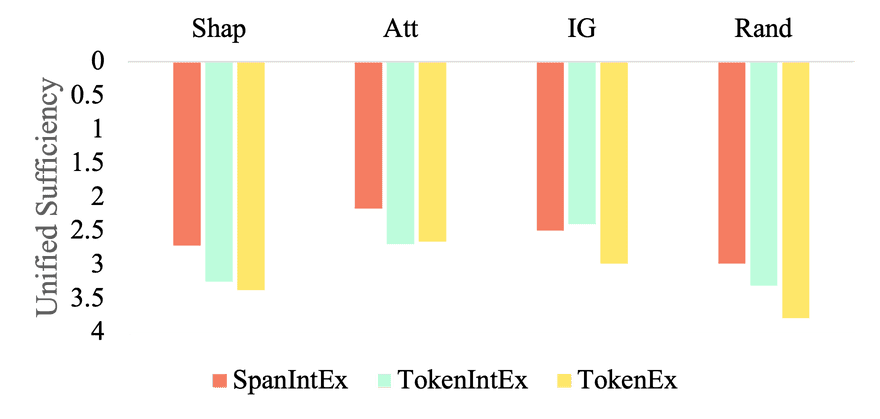}}
\subfigure[Sufficiency on FEVER dataset, the lower the better.]{\label{fig:subfig-suff-fever-bert:d}
\includegraphics[width=0.49\linewidth]{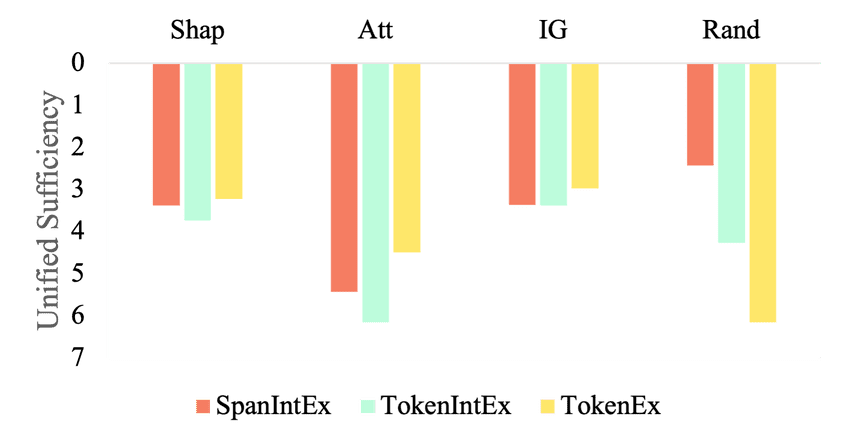}}
\caption{Unified Comprehensiveness and Sufficiency of three types of feature attribution explanations on SNLI and FEVER datasets using the BERT model. Subfigures (a) and (c) show Unified Comprehensiveness results, while (b) and (d) show Unified Sufficiency results. Explanations are generated by Shapley (Shap), Attention (Att), and Integrated Gradients (IG). Randomly selected span pairs, token pairs, and tokens are baselines corresponding to explanation type \spanintex, \tokenintex, and \tokenex\,  and form the group Random baseline (Rand). We set $k=3$ for top span interactions and adjust token counts as per \S\ref{sec:faithfulness:method}, also ensuring the random baseline matches the average token count of the top $k$ span interactions.}
\label{faithfulness-bert}
\scriptsize
\end{figure*}

\begin{figure*}[t]
\centering
\subfigure[Comprehensiveness on SNLI dataset, the higher the better.]{\label{fig:subfig-comp-snli-bart:a}
\includegraphics[width=0.49\linewidth]{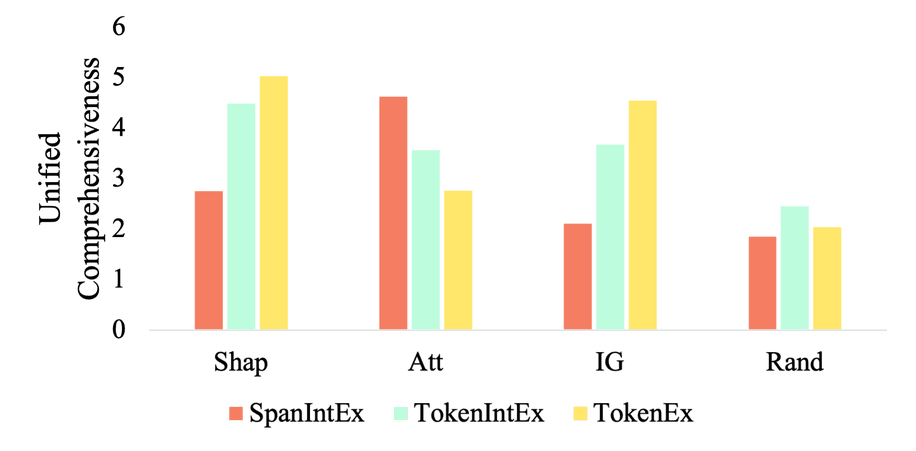}}
\subfigure[Comprehensiveness on FEVER dataset, the higher the better.]{\label{fig:subfig-comp-fever-bart:b}
\includegraphics[width=0.49\linewidth]{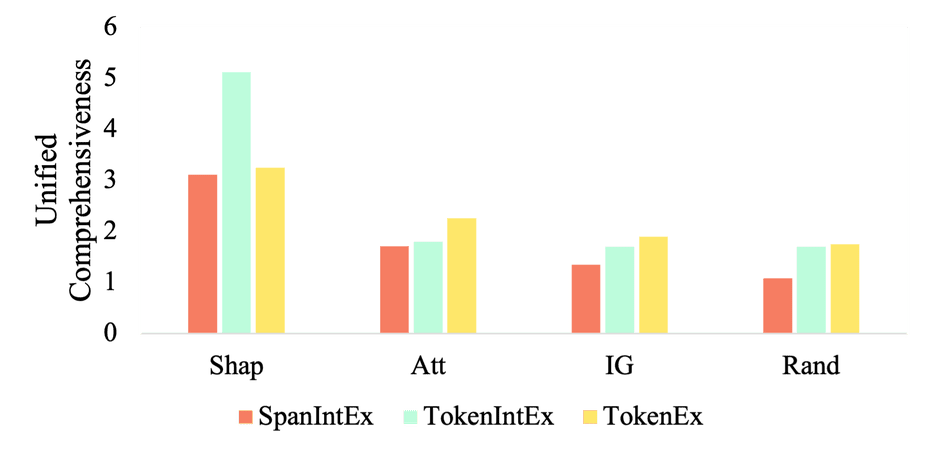}}
\subfigure[Sufficiency on SNLI dataset, the lower the better.]{\label{fig:subfig-suff-snli-bart:c}
\includegraphics[width=0.49\linewidth]{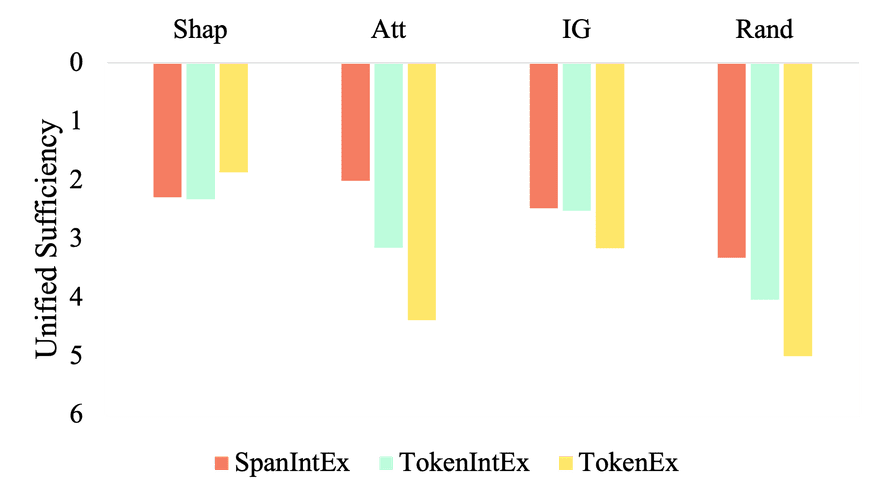}}
\subfigure[Sufficiency on FEVER dataset, the lower the better.]{\label{fig:subfig-suff-fever-bart:d}
\includegraphics[width=0.49\linewidth]{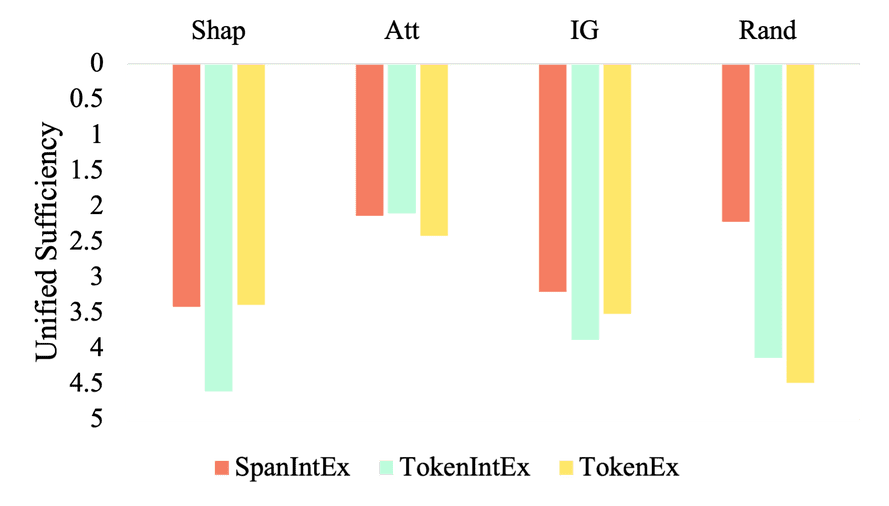}}
\caption{Unified Comprehensiveness and Sufficiency of three types of feature attribution explanations on SNLI and FEVER datasets using the BART model. Subfigures (a) and (c) show Unified Comprehensiveness results, while (b) and (d) show Unified Sufficiency results. Explanations are generated by Shapley (Shap), Attention (Att), and Integrated Gradients (IG). Randomly selected span pairs, token pairs, and tokens are baselines corresponding to explanation type \spanintex\, \tokenintex\, and \tokenex\,  and form the group Random baseline (Rand). We set $k=3$ for top span interactions and adjust token counts as per section \S\ref{sec:faithfulness:method}, ensuring the random baseline matches the average token count of the top $k$ span interactions.}
\label{faithfulness-bart}
\scriptsize
\end{figure*}

\subsection{Agreement with Human Annotation}

\begin{table}
\centering
\small
\begin{tabular}{lcccc}
\toprule
$\mathbf{E}$  & \textbf{Shap}  & \textbf{Att}   & \textbf{IG}    & \textbf{Rand}  \\ \midrule
              & \multicolumn{4}{c}{Interaction level agreement}                   \\
$\spanintex$  & \textbf{30.18} & \textbf{57.40} & \textbf{39.40} & \textbf{33.82} \\
$\tokenintex$ & 29.02          & 37.02          & 35.06          & 23.42          \\
$\tokenex$    & \textbf{-}     & \textbf{-}     & \textbf{-}     & \textbf{-}     \\
              & \multicolumn{4}{c}{Token level agreement}                         \\
$\spanintex$  & \textbf{75.63} & 78.26          & \textbf{76.52} & \textbf{76.96} \\
$\tokenintex$ & 74.89          & \textbf{79.60} & 73.19          & 74.96          \\
$\tokenex$    & 75.54          & 77.62          & 70.74          & 76.33          \\ \bottomrule
\end{tabular}%

\caption{Human Annotation Agreement Results (see \S\ref{sec:agreement:method}) on SNLI dataset when explanations are generated based on BERT. Interaction-level and Token-level agreement scores, Average Precision(\%), compared to human annotations for explanation types $\spanintex$, $\tokenintex$, $\tokenex$ generated by Shapley(\textbf{Shap}), Attention(\textbf{Att}), Integradiant Gradients(\textbf{IG}) respectively. Using the same attribution method, the highest alignment score for each category is highlighted in bold. \textbf{Rand} indicates the random baseline as described in \S\ref{sec:agreement:method}.}
\label{exper:agreement-bert-snli-f}
\end{table}

\begin{table}
\centering
\small
\begin{tabular}{lcccc}
\toprule
$\mathbf{E}$  & \textbf{Shap}  & \textbf{Att}   & \textbf{IG}    & \textbf{Rand}  \\ \midrule
              & \multicolumn{4}{c}{Interaction level agreement}                   \\
$\spanintex$  & \textbf{19.92} & \textbf{28.12} & \textbf{27.45} & \textbf{19.33} \\
$\tokenintex$ & 3.96           & 10.27          & 21.30 & 10.23          \\
$\tokenex$    & \textbf{-}     & \textbf{-}     & \textbf{-}     & \textbf{-}     \\
              & \multicolumn{4}{c}{Token level agreement}                         \\
$\spanintex$  & \textbf{66.95} & \textbf{68.71} & \textbf{72.24} & \textbf{67.5}  \\
$\tokenintex$ & 66.90          & 67.29          & 70.50          & 65.86          \\
$\tokenex$    & 58.07          & 56.88          & 61.07          & 63.10          \\ \bottomrule
\end{tabular}%

\caption{Human Annotation Agreement Results (see \S\ref{sec:agreement:method}) on the FEVER dataset when explanations are generated based on BERT. The rest of the settings are the same as Table \ref{exper:agreement-bert-snli-f}.}
\label{exper:agreement-bert-fever}
\end{table}

There is a notable gap between interaction-level and token-level agreement scores. For example, in Table \ref{exper:agreement-bert-snli-f}, the highest interaction-level agreement score for \spanintex\ explanations is 57.40\%, while the highest token-level agreement score for \spanintex\ is 78.26\%. A similar pattern is observed for \tokenintex. This suggests that although \spanintex\ and \tokenintex\ explanations align more with human reasoning than \tokenex\ explanations, pairing important tokens or spans into interactions that are plausible to humans remains challenging.

\begin{table}
\centering
\small
\begin{tabular}{lcccc}
\toprule
$\mathbf{E}$  & \textbf{Shap} & \textbf{Att}   & \textbf{IG} & \textbf{Rand}                \\ \midrule
              & \multicolumn{4}{c}{Interaction level agreement}                             \\
$\spanintex$  & \textbf{37.36} & \textbf{47.33} & \textbf{34.18} & {\color[HTML]{343434} \textbf{28.25}} \\
$\tokenintex$ & 32.36         & 35.17          & 33.06       & 13.80                        \\
$\tokenex$    & \textbf{-}    & \textbf{-}     & \textbf{-}  & \textbf{-}                   \\
              & \multicolumn{4}{c}{Token level agreement}                                   \\
$\spanintex$  & 80.16         & \textbf{76.28} & 82.76       & 70.04 \\
$\tokenintex$ & \textbf{84.9}  & 75.92          & \textbf{86.06} & \textbf{75.32} \\
$\tokenex$    & 65.74         & 73.71          & 70.44       & 73.34 \\ \bottomrule
\end{tabular}%

\caption{Human Annotation Agreement Results (see \S\ref{sec:agreement:method}) on SNLI dataset when explanations are generated based on BART. The rest settings are the same as Table \ref{exper:agreement-bert-snli-f}.}
\label{exper:agreement-bart-snli}
\end{table}

\begin{table}
\centering
\small
\begin{tabular}{lcccc}
\toprule
$\mathbf{E}$  & \textbf{Shap} & \textbf{Att} & \textbf{IG}    & \textbf{Rand}                \\ \midrule
              & \multicolumn{4}{c}{Interaction level agreement}                              \\
$\spanintex$ & \textbf{22.86} & \textbf{20.66} & \textbf{18.72} & {\color[HTML]{343434} \textbf{16.76}} \\
$\tokenintex$ & 4.71          & 2.64         & 10.54          & 8.64                         \\
$\tokenex$    & \textbf{-}    & \textbf{-}   & \textbf{-}     & \textbf{-}                   \\
              & \multicolumn{4}{c}{Token level agreement}                                    \\
$\spanintex$ & \textbf{68.77} & \textbf{65.33} & 70.22          & \textbf{69.51} \\
$\tokenintex$ & 67.01         & 63.40        & \textbf{70.98} & 68.11 \\
$\tokenex$    & 59.43         & 52.15        & 57.56          &  60.93 \\ \bottomrule
\end{tabular}%

\caption{Human Annotation Agreement Results (see \S\ref{sec:agreement:method}) on FEVER dataset when explanations are generated based on BART. The rest settings are the same as Table \ref{exper:agreement-bert-snli-f}.}
\label{exper:agreement-bart-fever}
\end{table}

\subsection{Simulatability}
Regarding insertion formats, for BERT models, text insertion ($I_{Text}$), which adds explanation text to the end of the input sequence, consistently outperforms symbol insertion ($I_{Sym}$), where symbols are added to the original input sequence, as shown in Tables \ref{exp:simulatability-form1-bert} and \ref{exp:simulatability-form2-bert}. However, the opposite effect is observed for BART models, as shown in Tables \ref{exp:simulatability-form1-bart} and \ref{exp:simulatability-form2-bart}. This indicates that simulatability results are sensitive to the explanation insertion form, highlighting the need for consistency in insertion form when comparing different explanation types.

\begin{table}
\centering
\small
\resizebox{\columnwidth}{!}{%
\begin{tabular}{llrrrrrr}
\toprule
\multirow{2}{*}{$\mathbf{D}$} & \multirow{2}{*}{$\mathbf{E}$} & \multicolumn{2}{c}{\textbf{Shap}} & \multicolumn{2}{c}{\textbf{Att}} & \multicolumn{2}{c}{\textbf{IG}} \\
                       &               & $\mathbf{SF}$ & $\mathbf{RSF}$ & $\mathbf{SF}$ & $\mathbf{RSF}$ & $\mathbf{SF}$ & $\mathbf{RSF}$ \\ \midrule
\multirow{2}{*}{SNLI}        & $\spanintex$                  & \textbf{87.9}    & \textbf{3.2}   & \textbf{86.7}   & \textbf{2.0}   & \textbf{88.9}   & \textbf{4.2}  \\
                       & $\tokenintex$ & 86.6          & 1.9            & 85.3          & 0.6            & 85.8          & 1.1            \\
                       & $\tokenex$    & 87.4          &2.7   & 85.7          & 1.0            & 86.0          & 1.3            \\ \midrule
\multirow{2}{*}{FEVER} & $\spanintex$  & 83.9          & 3.8            & \textbf{85.3} & \textbf{5.2}   & \textbf{85.2} & \textbf{5.1}   \\
                       & $\tokenintex$ & 82.7          & 2.6            & 84.0          & \textbf{3.9}   & 84.4          & 4.3            \\
                       & $\tokenex$    & \textbf{84.0} & \textbf{3.9}   & 82.3          & 2.2            & 81.8          & 1.7            \\ \bottomrule
\end{tabular}%
}
\caption{Simulatability results on SNLI and FEVER with BERT as the model used for all explanations $E\in{\spanintex, \tokenintex, \tokenex}$ generation with attribution method \textbf{Shapley}, \textbf{Attention}, and \textbf{Integrated Gradients} respectively. Note that insertion form $I_{Sym}$ is adopted for combining the explanations and the original input sequence for agent model $AM_{E}$, as depicted in \S\ref{sec:simulatability:method}. The agent models used for baseline $AM_O$, trained without explanations, have simulation F1 scores, as denoted in \S\ref{sec:simulatability:method}, of 84.7\%  and 80.0\% on test set shared with other agent models $AM_E$, as denoted in \S\ref{sec:simulatability:method}. We set $k=1$ for top \spanintex\ and calculated the number of top \tokenintex\ and \tokenex\ accordingly as stated in \S\ref{sec:faithfulness:method}. The largest increases are highlighted in bold for each dataset with the identical attribution method.}
\label{exp:simulatability-form1-bert}
\end{table}

\begin{table}
\centering
\small
\resizebox{\columnwidth}{!}{%
\begin{tabular}{llrrrrrr}
\toprule
\multirow{2}{*}{$\mathbf{D}$} & \multirow{2}{*}{$\mathbf{E}$} & \multicolumn{2}{c}{\textbf{Shap}} & \multicolumn{2}{c}{\textbf{Att}} & \multicolumn{2}{c}{\textbf{IG}} \\
                       &               & $\mathbf{SF}$ & $\mathbf{RSF}$ & $\mathbf{SF}$ & $\mathbf{RSF}$ & $\mathbf{SF}$ & $\mathbf{RSF}$ \\ \midrule
\multirow{2}{*}{SNLI}        & $\spanintex$                  & \textbf{87.8}    & \textbf{3.1}   & 87.1        & 2.4       & 88.2   & 3.5  \\
                       & $\tokenintex$ & 86.5          & 1.8            & \textbf{87.8} & \textbf{3.1}            & 86.4          & 1.7            \\
                       & $\tokenex$    & 87.0          & 2.3   & 86.3          & 1.6            & \textbf{88.4}          & \textbf{3.7}            \\ \midrule
\multirow{2}{*}{FEVER} & $\spanintex$  & 85.7          & 5.6            & 85.1 & 5.0            & \textbf{86.0} & \textbf{5.9}   \\
                       & $\tokenintex$ & 81.9          & 1.8            & \textbf{85.6} & 5.5            & 84.3          & 4.2            \\
                       & $\tokenex$    & \textbf{85.8} & \textbf{5.7}   & 84.5          & 4.4            & 82.0          & 1.9            \\ \bottomrule
\end{tabular}%

}
\caption{Simulatability results on SNLI and FEVER with BERT as the model used for all input feature explanation generation. Note that insertion form $I_{Text}$ is adopted for combining the explanations and the original input sequence for agent model $AM_{E}$, as depicted in \S\ref{sec:simulatability:method}.  The agent models used for baseline $AM_O$, which are trained without explanations, have simulation F1 scores of 84.7\% and 80.0\% on the test sets shared with agent model $AM_E$. The other setting is the same as Table \ref{exp:simulatability-form1-bert}}
\label{exp:simulatability-form2-bert}
\end{table}

\begin{table}
\centering
\small
\resizebox{\columnwidth}{!}{%
\begin{tabular}{llrrrrrr}
\toprule
\multirow{2}{*}{$\mathbf{D}$} & \multirow{2}{*}{$\mathbf{E}$} & \multicolumn{2}{c}{\textbf{Shap}} & \multicolumn{2}{c}{\textbf{Att}} & \multicolumn{2}{c}{\textbf{IG}} \\
                       &               & $\mathbf{SF}$ & $\mathbf{RSF}$ & $\mathbf{SF}$ & $\mathbf{RSF}$ & $\mathbf{SF}$ & $\mathbf{RSF}$ \\ \midrule
\multirow{2}{*}{SNLI}        & $\spanintex$                  & 87.8        & 7.9        & \textbf{87.3}   & \textbf{7.4}   & \textbf{86.5}   & \textbf{6.6}  \\
                       & $\tokenintex$ & 83.5          & 3.6            & 84.2          & 4.3            & 84.6          & 4.7            \\
                       & $\tokenex$    & \textbf{88.2} & \textbf{8.3}   & 81.4          & 1.5            & 85.8          & 5.9            \\ \midrule
\multirow{2}{*}{FEVER} & $\spanintex$  & \textbf{80.6} & 7.2            & \textbf{76.1}          & \textbf{2.7}            & \textbf{75.2} & \textbf{1.8}   \\
                       & $\tokenintex$ & 78.9          & 5.5            & 75.9 & 2.5            & 74.7          & 1.3            \\
                       & $\tokenex$    & 80.1          & \textbf{6.7}   & 75.0          & 1.6            & 74.1          & 0.7            \\ \bottomrule
\end{tabular}%

}
\caption{Simulatability results on SNLI and FEVER with BART as the model used for all input feature explanation generation. Note that insertion form $I_{Sym}$ is adopted for combining the explanations and the original input sequence for agent model $AM_{E}$, as depicted in \S\ref{sec:simulatability:method}. The base agent models $AM_O$ trained without explanations have the simulation f1 scores of 79.9\% and 73.4\%, respectively on the test sets sharing with other agent models $AM_E$. The other setting is the same as Table \ref{exp:simulatability-form1-bert}}
\label{exp:simulatability-form1-bart}
\end{table}

\begin{table}
\centering
\small
\resizebox{\columnwidth}{!}{%
\begin{tabular}{llrrrrrr}
\toprule
\multirow{2}{*}{$\mathbf{D}$} & \multirow{2}{*}{$\mathbf{E}$} & \multicolumn{2}{c}{\textbf{Shap}} & \multicolumn{2}{c}{\textbf{Att}} & \multicolumn{2}{c}{\textbf{IG}} \\
                       &               & $\mathbf{SF}$ & $\mathbf{RSF}$ & $\mathbf{SF}$ & $\mathbf{RSF}$ & $\mathbf{SF}$ & $\mathbf{RSF}$ \\ \midrule
\multirow{2}{*}{SNLI}        & $\spanintex$                  & \textbf{86.8}    & \textbf{6.9}   & \textbf{85.0}   & \textbf{5.1}   & 84.3       & \textbf{4.4}       \\
                       & $\tokenintex$ & 81.2          & 1.3            & 82.6          & 2.7            & \textbf{81.6} & 1.7            \\
                       & $\tokenex$    & 83.3          & 3.4            & 83.8          & 3.9            & 82.2          & 2.3            \\ \midrule
\multirow{2}{*}{FEVER} & $\spanintex$  & \textbf{78.2}          & \textbf{4.8}            & \textbf{75.3}          & \textbf{1.9}            & \textbf{74.6} & \textbf{1.2}   \\
                       & $\tokenintex$ & 74.8          & 1.4            & 74.1          & 0.7            & 73.9          & 0.5            \\
                       & $\tokenex$    & 77.6 & 4.2   & 74.9          & 1.5            & 73.6          & 0.3            \\ \bottomrule
\end{tabular}%

}
\caption{Simulatability results on SNLI and FEVER with BART as the model used for all input feature explanation generation respectively. Note that insertion form $I_{Text}$ is adopted for combining the explanations and the original input sequence for agent model $AM_{E}$, as depicted in \S\ref{sec:simulatability:method}. The base agent models $AM_O$ trained without explanations have the simulation f1 scores of 79.9\% and 73.4\%, respectively. The other setting is the same as Table \ref{exp:simulatability-form1-bert}}
\label{exp:simulatability-form2-bart}
\end{table}

\subsection{Complexity}

\begin{table}
\centering
\small
\resizebox{\columnwidth}{!}{%
\begin{tabular}{llccccc}
\toprule
\textbf{Dataset} & \textbf{$E$} & \textbf{Shapley} & \textbf{Attention} & \textbf{IG} & \textbf{R} & \textbf{U} \\ \hline
\multirow{2}{*}{SNLI}  & $\spanintex$  & 2.05          & \textbf{1.11} & 2.10          & 2.19 & 2.62 \\
                       & $\tokenintex$ & \textbf{1.72} & 1.43          & 2.30          & -    & -    \\
                       & $\tokenex$    & 2.08          & 1.64          & \textbf{1.91} & -    & -    \\ \midrule
\multirow{2}{*}{FEVER} & $\spanintex$  & 2.78          & \textbf{2.22} & 2.90          & 2.98 & 3.18 \\
                       & $\tokenintex$ & 3.12          & 3.07          & 3.15          & -    & -    \\
                       & $\tokenex$    & \textbf{2.76} & \textbf{2.22} & \textbf{2.57} & -    & -    \\ \bottomrule
\end{tabular}%
}
\caption{Complexity results on SNLI and FEVER datasets for three types of explanations generated by different attribution methods based on BERT model. The \textbf{Random} baseline represents the complexity score obtained by randomly generated scores in the range [0,1], ensuring the same number of scores as the number of explanations used. The \textbf{Upperbound} is calculated by setting all the attribution scores to the same value while ensuring the same number of scores as the number of explanations used. The lowest complexity score for each specific explanation type compared is highlighted in bold when the explanations are generated by each attribution method.}
\label{exp:complexity-bert}
\end{table}

\begin{table}
\centering
\small
\resizebox{\columnwidth}{!}{%
\begin{tabular}{llccccc}
\toprule
\textbf{Dataset}       & \textbf{$E$} & \textbf{Shapley} & \textbf{Attention} & \textbf{IG}   & \textbf{R} & \textbf{U} \\ \midrule
\multirow{2}{*}{SNLI}  & $\spanintex$ & 1.90             & \textbf{1.93}      & \textbf{1.63} & 2.36       & 2.76       \\
 & $\tokenintex$ & 2.50          & 2.53 & 2.13 & - & - \\
 & $\tokenex$    & \textbf{1.86} & 1.95 & 1.94 & - & - \\ \midrule
\multirow{2}{*}{FEVER} & $\spanintex$ & \textbf{2.73}    & \textbf{3.03}      & \textbf{2.38} & 3.13       & 3.38       \\
 & $\tokenintex$ & 3.30          & 3.36 & 2.93 & - & - \\
 & $\tokenex$    & 2.82          & 3.07 & 3.19 & - & - \\ \bottomrule
\end{tabular}%
}
\caption{Complexity results on SNLI and FEVER datasets for three types of explanations generated by different attribution methods based on BART model. The other settings are the same as Table \ref{exp:complexity-bert}.}
\label{exp:complexity-bart}
\end{table}

\end{document}